\documentclass[pdflatex,sn-mathphys-num]{sn-jnl}

\usepackage{graphicx}%
\usepackage{multirow}%
\usepackage{amsmath,amssymb,amsfonts}%
\usepackage{bbm}
\usepackage{amsthm}%
\usepackage{mathrsfs}%
\usepackage[title]{appendix}%
\usepackage{xcolor}%
\usepackage{textcomp}%
\usepackage{manyfoot}%
\usepackage{booktabs}%
\usepackage{algorithm}%
\usepackage{algorithmicx}%
\usepackage{algpseudocode}%
\usepackage{listings}%

\theoremstyle{thmstyleone}%
\theoremstyle{thmstyletwo}%

\theoremstyle{thmstylethree}%

\begin{document}

\title[A Causal Machine Learning Framework for Treatment Personalization in Clinical Trials: Application to Ulcerative Colitis]{A Causal Machine Learning Framework for Treatment Personalization in Clinical Trials: Application to Ulcerative Colitis}


\author*[1]{\fnm{Cristian} \sur{Minoccheri}}\email{minoc@umich.edu}

\author[1]{\fnm{Sophia} \sur{Tesic}}

\author[1,3,4]{\fnm{Kayvan} \sur{Najarian}}

\author[1,2]{\fnm{Ryan} \sur{Stidham}}

\affil[1]{\orgdiv{Gilbert S. Omenn Department of Computational Medicine and Bioinformatics},  \orgname{University of Michigan}, \orgaddress{\street{1109 Geddes Avenue}, \city{Ann Arbor}, \postcode{48104}, \state{MI}, \country{USA}}}

\affil[2]{\orgdiv{Department of Gastroenterology}, \orgname{University of Michigan}, \orgaddress{\street{1109 Geddes Avenue}, \city{Ann Arbor}, \postcode{48104}, \state{MI}, \country{USA}}}

\affil[3]{\orgdiv{Department of Emergency Medicine}, \orgname{University of Michigan}, \orgaddress{\street{1109 Geddes Avenue}, \city{Ann Arbor}, \postcode{48104}, \state{MI}, \country{USA}}}

\affil[4]{\orgdiv{Department of Electrical Engineering and Computer Science}, \orgname{University of Michigan}, \orgaddress{\street{1109 Geddes Avenue}, \city{Ann Arbor}, \postcode{48104}, \state{MI}, \country{USA}}}

\abstract{Randomized controlled trials estimate average treatment effects, but treatment response heterogeneity motivates personalized approaches. A critical question is whether statistically detectable heterogeneity translates into improved treatment decisions -- these are distinct questions that can yield contradictory answers. We present a modular causal machine learning framework that evaluates each question separately: permutation importance identifies which features predict heterogeneity, best linear predictor (BLP) testing assesses statistical significance, and doubly robust policy evaluation measures whether acting on the heterogeneity improves patient outcomes. We apply this framework to patient-level data from the UNIFI maintenance trial of ustekinumab in ulcerative colitis, comparing placebo (n=144), standard-dose ustekinumab every 12 weeks (n=143), and dose-intensified ustekinumab every 8 weeks (n=274), using cross-fitted X-learner models with baseline demographics, medication history, week-8 clinical scores, laboratory biomarkers, and video-derived endoscopic features. BLP testing identified strong associations between endoscopic features and treatment effect heterogeneity for ustekinumab versus placebo (p\textless{}0.001), yet doubly robust policy evaluation showed no improvement in expected remission from incorporating endoscopic features (95\% CI: $-$1.6 to +6.6 pp), and out-of-fold multi-arm evaluation showed worse performance (30.5\% vs 36.8\% remission). Diagnostic comparison of prognostic contribution against policy value revealed that endoscopic scores behaved as disease severity markers -- improving outcome prediction in untreated patients but adding noise to treatment selection -- while clinical variables (fecal calprotectin, age, CRP) captured the decision-relevant variation. These results demonstrate that causal machine learning applications to clinical trials should include policy-level evaluation alongside heterogeneity testing: the gap between statistical significance and decision performance is itself informative and can help distinguish prognostic features from genuine effect modifiers.}

\maketitle

\section{Introduction}\label{sec1}

Heterogeneity in treatment response is a persistent challenge across therapeutic areas. Randomized controlled trials (RCTs) provide unbiased estimates of average treatment effects, but outcomes vary considerably across patients \cite{vanish2025}, motivating principled frameworks for identifying subgroups with differential treatment benefit. A critical question is whether estimated heterogeneity translates into improved treatment decisions, rather than reflecting statistical separation alone.

While methods for estimating heterogeneous treatment effects have advanced substantially in the statistical and econometric literature, their systematic application to clinical trial data remains limited. While prognostic models cannot distinguish patients who would improve on their own from those who improve because of treatment, heterogeneous treatment effects estimate the causal effect of treatment for each individual patient, directly informing which patients benefit enough from a specific therapy to justify its costs and risks. Heterogeneous treatment effects can be estimated through conditional average treatment effects (CATEs), the expected treatment effect conditional on a patient's covariates. Foundational work on causal forests \cite{athey2018} and meta-learners \cite{kunzel2019} has established flexible approaches for estimating CATEs, and doubly robust estimation \cite{robins1994,bang2005,kennedy2023} provides a principled basis for evaluating individualized treatment rules. A parallel literature on policy learning has formalized the distinction between detecting effect heterogeneity and learning treatment policies that improve expected outcomes \cite{athey2021,kitagawa2018}. Best linear predictor testing \cite{chernozhukov2018} offers a framework for assessing whether specific feature groups contribute to treatment effect heterogeneity beyond other covariates. Despite these methodological advances, few studies have applied these tools in an integrated manner to clinical trial data, and important practical questions remain: how to distinguish prognostic features from genuine effect modifiers, whether in-sample heterogeneity detection translates to improved out-of-sample treatment assignment, and how to handle multi-arm comparisons that arise naturally in dose-finding and maintenance trials.

Recent work \cite{vanish2025} applied causal forests to RCT data with a focus on identifying interpretable subgroup splits and demonstrated the value of tree-based effect estimation. The present work differs in emphasis and scope: rather than focusing on subgroup identification, we combine X-learner estimation with doubly robust policy evaluation to directly assess whether candidate feature sets improve treatment-assignment decisions, and we extend the analysis to multi-arm settings with nested out-of-fold evaluation. This policy-oriented perspective -- evaluating whether heterogeneity detection translates into better decisions -- addresses a gap that is underexplored in the clinical trial literature.

We present a causal machine learning framework for treatment personalization in RCTs that combines X-learner CATE estimation, permutation-based feature importance, best linear predictor (BLP) testing \cite{chernozhukov2018}, doubly robust policy value evaluation, and prognostic performance modeling. This framework is designed to separate statistical associations from improvements in treatment selection, and includes out-of-fold evaluation to quantify generalization of learned treatment rules. The methodology is applicable across therapeutic areas where treatment effect heterogeneity is of interest. It helps identify which features predict heterogeneity, whether the heterogeneity is statistically significant, and whether it improves treatment decisions.

To demonstrate the framework, we apply it to individual-patient data from the UNIFI randomized maintenance trial of ustekinumab in ulcerative colitis (UC) \cite{sands2019}. Treatment personalization in inflammatory bowel disease is an area of growing interest \cite{ceccato2025,gisbert2020}, but most existing work focuses on prognostic prediction -- identifying who will respond to a given therapy -- rather than treatment effect modification, which asks who benefits more from one therapy versus another. This distinction has direct implications for clinical decision-making, yet it has not been formally evaluated in UC maintenance trials using causal inference methods.

Endoscopic measures such as the Mayo Endoscopic Score (MES) are widely used to assess mucosal healing \cite{schroeder1987,travis2013}, and early endoscopic response has been studied as a prognostic predictor of long-term remission \cite{shah2016}. However, whether endoscopic features serve as stable effect modifiers -- helping to identify patients who require dose intensification (every 8 weeks versus every 12 weeks), those who benefit from continued active therapy versus placebo withdrawal, or those who would benefit from a different maintenance strategy altogether -- remains unclear given inter-rater variability, invasiveness, and cost \cite{harbord2017}.

We demonstrate the framework using individual-patient data from the UNIFI maintenance trial, evaluating whether endoscopic features improve personalized treatment assignment beyond routinely available clinical markers across binary and multi-arm comparisons.
The code used for this study is available at \url{https://github.com/Minoch/UNIFI_causal}.

The main contributions are as follows:

1) A modular causal machine learning framework for treatment personalization in RCTs that integrates CATE estimation, feature importance, statistical testing, and doubly robust policy evaluation into a single pipeline -- designed so that each component answers a different question: which features predict heterogeneity (permutation importance), is the heterogeneity statistically significant (BLP testing), and does it improve treatment decisions (policy value).

2) Empirical demonstration, using the UNIFI maintenance trial in ulcerative colitis, that these questions can yield contradictory answers: endoscopic features showed strong statistical associations with treatment effect heterogeneity (BLP z = 8.28) yet produced no improvement in treatment assignment -- and degraded out-of-fold policy value in the multi-arm setting (30.5\% vs. 36.8\% remission). This underscores that heterogeneity detection and decision performance must be evaluated separately.

3) A diagnostic strategy for distinguishing prognostic features from effect modifiers by comparing their contribution to outcome prediction (incremental Brier score) against their contribution to policy value across treatment arms. Applied here, this approach identified endoscopic scores as severity markers -- improving prognosis in untreated patients but adding noise to treatment selection -- while routine clinical variables (calprotectin, age, CRP) captured the decision-relevant variation.

\section{Dataset}\label{sec2}

This analysis utilizes data from the UNIFI maintenance-phase randomized controlled trial of ustekinumab in patients with moderate-to-severe ulcerative colitis \cite{sands2019}. The cohort consists of the intention-to-treat (ITT) population: patients who received induction therapy and were randomized to maintenance regimens of placebo, ustekinumab every 12 weeks (Q12), or ustekinumab every 8 weeks (Q8) over 44 weeks \cite{sands2019,afif2024}.

The primary outcome was clinical remission at week 44 based on the full Mayo score, with non-completers classified as non-responders (intention-to-treat). Treatment assignment was determined by maintenance regimen indicators (MaintMed\_UST\_12 for Q12, MaintMed\_UST\_Q8 for Q8), with remaining subjects classified as placebo. Receipt of subcutaneous ustekinumab during induction (InductionMed\_UST\_SC) was included as a covariate. Treatment assignment variables defined comparison groups but were excluded from outcome models to avoid information leakage.

The feature set for modeling included patient age and sex, binary medication history variables (no prior biologic exposure, baseline immunomodulator use, history of treatment intolerance or refractoriness), and intention-to-treat inclusion indicator. Week-8 laboratory measurements provided objective markers of inflammation, including fecal calprotectin (CALPL\_wk8) and C-reactive protein (CRPL\_wk8).

Clinical activity variables were derived from week-8 assessments: absolute bowel movement count (ABSSTOOL\_wk8), rectal bleeding score, stool frequency score, physician global assessment, and partial Mayo score (PMAYO\_wk8). Change scores for each of these symptom-based measures were computed from baseline to week 8 (e.g., pMayo\_abs\_0\_8, RBscore\_abs\_0\_8, SFscore\_abs\_0\_8, PGscore\_abs\_0\_8, ABSstool\_abs\_0\_8).

Endoscopic and video-derived features assessed mucosal inflammation and healing via detailed segmental scoring of the sigmoid colon. For each colonoscopy at week 8, the percent of tissue displaying each Mayo endoscopic subscore (0–3) was quantified (MES\_0\_perc\_wk8, MES\_1\_perc\_wk8, MES\_2\_perc\_wk8, MES\_3\_perc\_wk8), along with the recently introduced Cumulative Disease Score (CDS) \cite{stidham2024}. Changes in these distributions from baseline to week 8 were also included to capture early response dynamics (MES\_0\_perc\_Abs\_0\_8, MES\_1\_perc\_Abs\_0\_8, MES\_2\_perc\_Abs\_0\_8, MES\_3\_perc\_Abs\_0\_8, CDS\_abs\_0\_8).

All feature definitions are listed in Table~\ref{tab:analysis_feature_groups} below. Missing data were handled using median imputation for continuous variables and mode imputation for binary indicators. Continuous features were standardized with z-score normalization, while binary variables retained their original scale.

\begin{table}[h]
\centering
\caption{Dataset Feature Descriptions}
\label{tab:analysis_feature_groups}
\begin{scriptsize}
\begin{tabular}{@{}ll@{}}
\toprule
\textbf{Feature} & \textbf{Description} \\
\midrule
\multicolumn{2}{l}{\textit{Endoscopic Features}} \\
MES\_0\_perc\_wk8 & Percent of sigmoid with Mayo 0 at week 8 \\
MES\_1\_perc\_wk8 & Percent of sigmoid with Mayo 1 at week 8 \\
MES\_2\_perc\_wk8 & Percent of sigmoid with Mayo 2 at week 8\\
MES\_3\_perc\_wk8 & Percent of sigmoid with Mayo 3 at week 8 \\
MES\_0\_perc\_Abs\_0\_8 & Change in percent Mayo 0 in sigmoid, week 0 to 8 \\
MES\_1\_perc\_Abs\_0\_8 & Change in percent Mayo 1 in sigmoid, week 0 to 8 \\
MES\_2\_perc\_Abs\_0\_8 & Change in percent Mayo 2 in sigmoid, week 0 to 8 \\
MES\_3\_perc\_Abs\_0\_8 & Change in percent Mayo 3 in sigmoid, week 0 to 8 \\
CDS\_wk8 & CDS score for sigmoid at week 8  \\ & (continuous mucosal activity marker) \\
CDS\_abs\_0\_8 & Absolute change in CDS score, week 0 to 8 \\
\midrule
\multicolumn{2}{l}{\textit{Non-Endoscopic Features}} \\
AGE & Age in years at enrollment \\
SEX\_M & Binary indicator for male sex \\
BIONAIVE & No prior biologic exposure (medication history) \\
BIMM & Immunomodulator use at baseline (medication history) \\
RDIALL & History of intolerance or refractoriness  \\ & to immunomodulators/steroids \\
RDICORT & History of intolerance or refractoriness  \\ & to corticosteroids only \\
CALPL\_wk8 & Fecal calprotectin at week 8 (lab, gut inflammation) \\
CRPL\_wk8 & C-reactive protein at week 8 (lab, systemic inflammation) \\
ABSSTOOL\_wk8 & Absolute bowel movement count at week 8 \\
PGSCORE\_wk8 & Physician global assessment (0-3), week 8 \\
PMAYO\_wk8 & Partial Mayo score at week 8 (symptoms composite) \\
RBSCORE\_wk8 & Rectal bleeding score (0-3), week 8 \\
SFSCORE\_wk8 & Stool frequency score (0-3), week 8 \\
pMayo\_abs\_0\_8 & Change in partial Mayo score (week 0 to 8) \\
RBscore\_abs\_0\_8 & Change in rectal bleeding score (week 0 to 8) \\
SFscore\_abs\_0\_8 & Change in stool frequency score (week 0 to 8) \\
PGscore\_abs\_0\_8 & Change in physician global assessment (week 0 to 8) \\
ABSstool\_abs\_0\_8 & Change in absolute stool count (week 0 to 8) \\
InductionMed\_UST\_SC & Received subcutaneous UST induction (binary) \\
\midrule
\multicolumn{2}{l}{\textit{Treatment Variables}} \\

MaintMed\_UST\_12 & Maintenance assignment: UST every 12 weeks (binary) \\
MaintMed\_UST\_Q8 & Maintenance assignment: UST every 8 weeks (binary) \\
\midrule
\multicolumn{2}{l}{\textit{Primary Outcome}} \\
Remission\_Full\_CALC\_wk44\_ITT & Clinical remission at week 44 \\ & (full Mayo score, non-completers are non-responders) \\
\bottomrule
\end{tabular}
\end{scriptsize}
\end{table}

\section{Methods}

We employed the X-learner methodology \cite{kunzel2019} to estimate conditional average treatment effects (CATEs) and identify patient-level predictors of heterogeneous treatment response.

We conducted three analyses of increasing complexity: (1) combined ustekinumab maintenance (Q8 and Q12) versus placebo withdrawal, (2) dose-intensified (Q8) versus standard (Q12) maintenance, and (3) a multi-arm policy assigning each patient to the optimal strategy among all three options. The multi-arm analysis used nested cross-validation for unbiased out-of-fold policy value estimation.

The X-learner proceeds in three stages \cite{kunzel2019}. In the first stage, we fit separate outcome models for each treatment arm using gradient-boosted regression trees (XGBoost) to predict week-44 remission from baseline and week-8 features. For each binary comparison, models $\mu_0(X)$ and $\mu_1(X)$ were fit for control and treated patients respectively (800 trees, maximum depth 4, subsampling rate 0.8, column sampling rate 0.8, learning rate 0.05, L2 regularization 1.0). All models were cross-fitted using five-fold cross-validation.

In the second stage, we computed imputed treatment effects by generating pseudo-outcomes. For treated patients, $D_i^{(1)} = Y_i - \mu_0(X_i)$; for control patients, $D_i^{(0)} = \mu_1(X_i) - Y_i$. Separate CATE models were fit by regressing these pseudo-outcomes on baseline and week-8 features using gradient-boosted trees (400 estimators, maximum depth 4, subsampling rate 0.8, column sampling rate 0.8).

In the third stage, the two CATE estimates were combined using propensity score weighting. Since treatment was randomized, we used constant propensity scores equal to the empirical treatment probabilities. The final CATE estimate was:
\begin{equation}
\tau(X_i) = e \cdot \tau^{(0)}(X_i) + (1 - e) \cdot \tau^{(1)}(X_i),
\end{equation}
where $e = P(T=1)$ is the constant propensity score and $\tau^{(0)}(X_i)$ and $\tau^{(1)}(X_i)$ are the CATE estimates from control and treated groups respectively.

For the multi-arm comparison, we fit three outcome models $\mu_0(X)$, $\mu_{12}(X)$, and $\mu_8(X)$ using the same XGBoost architecture. The multi-arm policy was defined as:
\begin{equation}
\pi(X_i) = \arg\max_{t \in \{0, 12, 8\}} \mu_t(X_i).
\end{equation}

This policy was evaluated using doubly robust estimation with empirical randomization probabilities $P(T=0) = 0.257$, $P(T=\text{Q12}) = 0.255$, and $P(T=\text{Q8}) = 0.488$. To avoid overfitting, we employed nested five-fold stratified cross-validation: outcome models were fit on training folds and policies evaluated on held-out folds, with doubly robust contributions aggregated across all folds.

To quantify the relative contribution of endoscopic versus clinical features, we performed permutation-based feature importance analysis \cite{breiman2001,fisher2019} on the learned CATE models. For binary comparisons, we evaluated the second-stage model $\tau^{(1)}(X)$ fit on treated patients. For each feature, values were randomly permuted across patients 10 times and the increase in mean squared error relative to the unpermuted baseline was measured \cite{fisher2019}. The entire procedure was repeated 10 times to compute means and standard errors.

For the multi-arm analysis, we computed pseudo-outcomes representing the optimality gap -- the improvement achievable under the best alternative treatment:
\begin{equation}
D_i^{\text{gap}} = \max_{t \in \{0, 12, 8\}} \mu_t(X_i) - \mu_{T_i}(X_i).
\end{equation}
A gradient-boosted model (400 trees, maximum depth 4, subsampling rate 0.8, column sampling rate 0.8) was fit to predict this gap, and permutation importance was evaluated on this model.

Group-level importance was computed by summing individual feature importances within two categories: (1) endoscopic features (MES percentages, MES percentage changes, and CDS scores), and (2) non-endoscopic features (demographics, medication history, week-8 clinical scores, and laboratory biomarkers).

To formally test whether endoscopic features contribute beyond clinical variables, we applied the best linear predictor framework \cite{chernozhukov2018}. For each feature group (clinical and endoscopic), we computed two summary statistics per patient (mean and standard deviation across features within that group) and regressed the estimated CATEs $\tau(X_i)$ on the concatenated summaries using ordinary least squares. Statistical inference was performed via multiplier bootstrap resampling with 1000 iterations. We constructed a Wald-type test statistic by summing the squared z-scores for the two endoscopic coefficients.

To assess the practical utility of endoscopic features for treatment selection, we evaluated hypothetical treatment assignment policies using doubly robust estimation \cite{kennedy2023,robins1994,bang2005}. For each feature set (all features or clinical only), we defined a policy $\pi(X_i) = \arg\max_{t} \mu_t(X_i)$ assigning the treatment with the highest predicted outcome. The policy value was estimated as:
\begin{equation}
V(\pi) = \frac{1}{n} \sum_{i=1}^{n} \left[ \mu_{\pi(X_i)}(X_i) + \frac{\mathbbm{1}(T_i = \pi(X_i))}{e_{\pi(X_i)}} \left( Y_i - \mu_{\pi(X_i)}(X_i) \right) \right],
\end{equation}
where $\mathbbm{1}(T_i = \pi(X_i))$ is an indicator that the observed treatment matches the policy assignment, $e_{\pi(X_i)}$ is the probability of receiving treatment $\pi(X_i)$ under randomization, and $\mu_t(X_i)$ are the outcome models from the first stage of the X-learner. This estimator is consistent if either the outcome models or the propensity scores are correctly specified \cite{robins1994,bang2005}.

We compared policy values between the endoscopy-informed policy $\pi_{\text{all}}$ and the clinical-only policy $\pi_{\text{clinical}}$, computing $\Delta = V(\pi_{\text{all}}) - V(\pi_{\text{clinical}})$. Confidence intervals were constructed using paired bootstrap resampling with 5000 iterations, accounting for the paired nature of the comparison. For the multi-arm policy, out-of-fold estimates from nested cross-validation provided a conservative assessment of generalization.

As a complementary analysis, we evaluated the predictive accuracy of baseline and week-8 features for week-44 remission using supervised classification, independent of treatment assignment. Calibrated XGBoost classifiers (600 trees, maximum depth 4, subsampling rate 0.85, column sampling rate 0.85, learning rate 0.05, L2 regularization 1.0) were evaluated via five-fold stratified cross-validation with isotonic regression calibration. Performance was assessed using AUROC, and Brier score. We compared models with all features versus clinical features only, both full cohort and stratified by treatment arm, using paired bootstrap resampling (5000 iterations) to construct 95\% confidence intervals for metric differences. The full cohort comprises all patients across treatment arms (n=561); arm-stratified analyses assess consistency within each group.

We also computed an incremental Brier score $\text{Brier}_{\text{incremental}}$ statistic to quantify the additional proportion of null-model error explained by adding endoscopic features beyond clinical features:
\begin{equation}
\text{Brier}_{\text{incremental}} = \frac{\text{Brier}_{\text{clinical}} - \text{Brier}_{\text{all}}}{\text{Brier}_{\text{null}}},
\end{equation}
where $\text{Brier}_{\text{null}}$ is the Brier score of a null model predicting the marginal outcome probability (i.e., $\hat{p}_i = \bar{Y}$ for all patients). This metric is equivalent to the difference in Index of Prediction Accuracy (IPA) \cite{kattan2018} between the two models:
\begin{equation}
\text{Brier}_{\text{incremental}} = \text{IPA}_{\text{all}} - \text{IPA}_{\text{clinical}},
\end{equation}
where $\text{IPA} = 1 - \text{Brier}/\text{Brier}_{\text{null}}$. Positive values indicate that endoscopic features capture additional predictive signal, while negative values suggest that endoscopic features add noise or cause overfitting.

To validate the CATE importance findings using observed treatment effects, we computed subgroup average treatment effects (ATEs) within pre-specified patient subgroups. Because treatment was randomized, the ATE within any subgroup is identified by a simple difference in mean outcomes:
\begin{equation}
\widehat{\text{ATE}}_S = \frac{1}{n_{1,S}} \sum_{i \in S: T_i=1} Y_i - \frac{1}{n_{0,S}} \sum_{i \in S: T_i=0} Y_i,
\end{equation}
where $S$ denotes the subgroup, $n_{1,S}$ and $n_{0,S}$ are the numbers of treated and control patients in the subgroup, and $Y_i$ is the binary remission outcome. Subgroups were defined by the top features from permutation importance analysis, using clinically established thresholds where available: fecal calprotectin $\geq$150 $\mu$g/g \cite{turner2021}, CRP $\geq$5 mg/L \cite{vermeire2006}, and partial Mayo score $\geq$3 \cite{lewis2008}. Age used median split as no established clinical cutoff exists. Confidence intervals were constructed using bootstrap resampling with 5000 iterations. This analysis provides model-free validation: if CATE importance rankings reflect true heterogeneity, then subgroups defined by high-importance features should exhibit meaningfully different observed ATEs.

For the reader's convenience, we briefly summarize the methods in Table~\ref{tab:methods_interpretation}.

\begin{table}[ht]
\centering
\caption{Plain-Language Summary of the Methods}
\label{tab:methods_interpretation}
\begin{small}
\begin{tabular}{@{}p{0.28\textwidth}p{0.65\textwidth}@{}}
\hline
\textbf{Method} & \textbf{Summary} \\
\hline
X-learner CATE estimation &
Builds separate prediction models for treated and untreated patients, then estimates each patient's personal treatment benefit by asking: ``What would this patient's outcome have been under the other treatment?'' \\[4pt]

Permutation importance &
Identifies which patient characteristics matter most for treatment benefit by shuffling each feature and measuring how much worse the model becomes at predicting who benefits. \\[4pt]

BLP testing &
Tests whether endoscopic features explain variation in estimated treatment effects beyond what clinical features already capture. A significant result indicates association, not necessarily improved decisions. \\[4pt]

Doubly robust policy evaluation &
Simulates what would happen if a model's treatment recommendations were followed, estimating the resulting remission rate. Answers whether detecting heterogeneity actually translates into better treatment assignments. \\[4pt]

Incremental Brier score &
Measures whether adding endoscopic features improves prediction of who will achieve remission. \\[4pt]

Subgroup ATEs &
A model-free check; patients are split into clinically defined groups and observed remission rates are compared between treatment and control within each group, using no modeling. \\
\hline
\end{tabular}
\end{small}
\end{table}

\section{Results}

\begin{figure}[ht]
\centering
\includegraphics[width=0.9\textwidth]{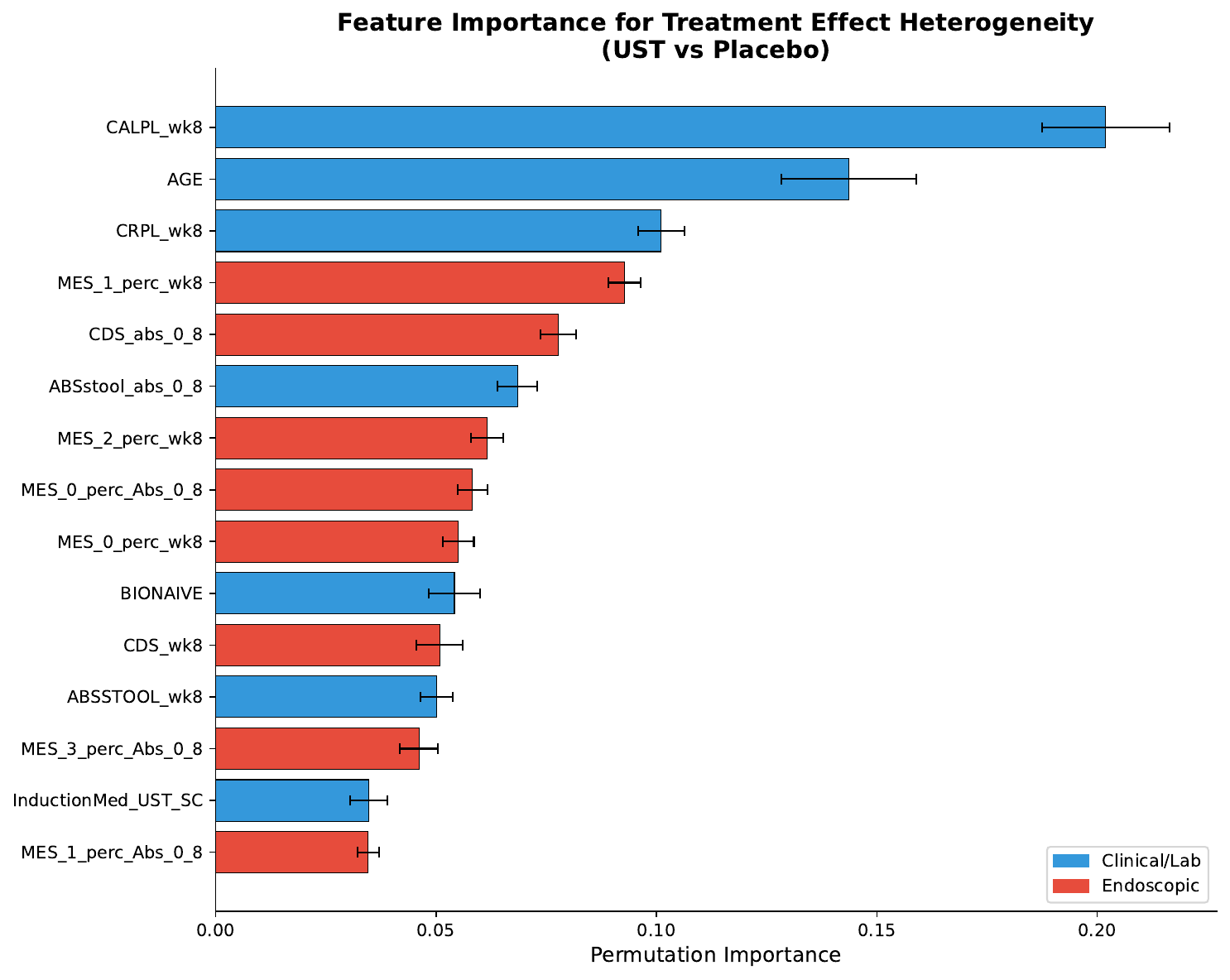}
\caption{Permutation importance for predicting heterogeneous treatment effects (UST vs Placebo). Features are color-coded by category: blue indicates clinical/laboratory features, red indicates endoscopic features. Error bars represent 95\% confidence intervals. Clinical features, particularly inflammatory biomarkers (fecal calprotectin, CRP) and patient age, show the highest importance for identifying treatment effect heterogeneity.}
\label{fig:feature_importance}
\end{figure}

Permutation-based feature importance revealed that clinical and demographic variables consistently showed greater predictive value than endoscopic features for identifying differential treatment response (Table~\ref{tab:importance_overview}, Figure~\ref{fig:feature_importance}).

For the comparison of combined ustekinumab maintenance therapy versus placebo withdrawal following successful induction, non-endoscopic features achieved aggregate importance of 0.79, approximately 1.5-fold higher than endoscopic video analysis features (importance = 0.54). Among individual predictors, the most informative clinical variables were fecal calprotectin at week 8 (importance = 0.202), age (0.144), C-reactive protein at week 8 (0.101), and percentage of sigmoid with MES 1 at week 8 (0.093). Within endoscopic features, change in CDS from baseline showed the highest importance (0.078), followed by change in MES 0 percentage from baseline (0.058).

The analysis comparing dose-intensified maintenance (ustekinumab every 8 weeks) versus standard dosing (every 12 weeks) among patients receiving active therapy revealed similar patterns. Non-endoscopic features demonstrated aggregate importance of 0.71, approximately 1.4-fold higher than endoscopic features (importance = 0.50). The most predictive clinical variables for dosing decisions were fecal calprotectin (importance = 0.133), age (0.129), C-reactive protein (0.096), and change in MES 1 percentage from baseline (0.093).

For the multi-arm policy that assigns each patient to the optimal maintenance strategy among placebo, Q12, or Q8, the importance ratio decreased to 1.2-fold (non-endoscopic: 0.92 $\pm$ 0.02, endoscopic: 0.75 $\pm$ 0.01). Although this suggests that endoscopic features play a relatively larger role in complex multi-arm treatment decisions, the absolute magnitude of endoscopic importance remained lower than that of clinical features.

\begin{table}[ht]
\centering
\caption{Aggregate Feature Importance Comparing Endoscopic vs. Clinical Features}
\begin{tabular}{lccc}
\hline
Comparison & Non-Endoscopic & Endoscopic & Ratio \\
\hline
UST vs Placebo & 0.79 $\pm$ 0.02 & 0.54 $\pm$ 0.01 & 1.46$\times$ \\
Q8 vs Q12 & 0.71 $\pm$ 0.03 & 0.50 $\pm$ 0.02 & 1.42$\times$ \\
Multi-arm policy & 0.92 $\pm$ 0.02 & 0.75 $\pm$ 0.01 & 1.22$\times$ \\
\hline
\end{tabular}
\label{tab:importance_overview}
\end{table}

To formally assess whether endoscopic features contribute beyond clinical variables, we applied BLP testing \cite{chernozhukov2018}. For ustekinumab versus placebo, the BLP test was strongly significant (z = 8.28, p \textless{} 0.001), indicating that endoscopic summaries explain additional variation in estimated treatment effects beyond clinical summaries (Table~\ref{tab:blp_results}) \cite{travis2013,shah2016}.

However, for dose intensification (Q8 vs Q12), the test was non-significant (z = 1.01, p = 0.311), and the multi-arm policy showed only borderline significance (z = 1.82, p = 0.069). Thus, while endoscopic features show statistical significance for biologic continuation versus withdrawal, they contribute minimally to dosing optimization or multi-arm assignment.

\begin{table}[ht]
\centering
\caption{Best Linear Predictor Test Results for Endoscopic Contribution Beyond Clinical Features}
\begin{tabular}{lcc}
\hline
Comparison & Z-statistic & P-value \\
\hline
UST vs Placebo & 8.28 & $<$0.001 \\
Q8 vs Q12 & 1.01 & 0.311 \\
Multi-arm policy & 1.82 & 0.069 \\
\hline
\end{tabular}
\label{tab:blp_results}
\end{table}

\begin{figure}[ht]
\centering
\includegraphics[width=0.9\textwidth]{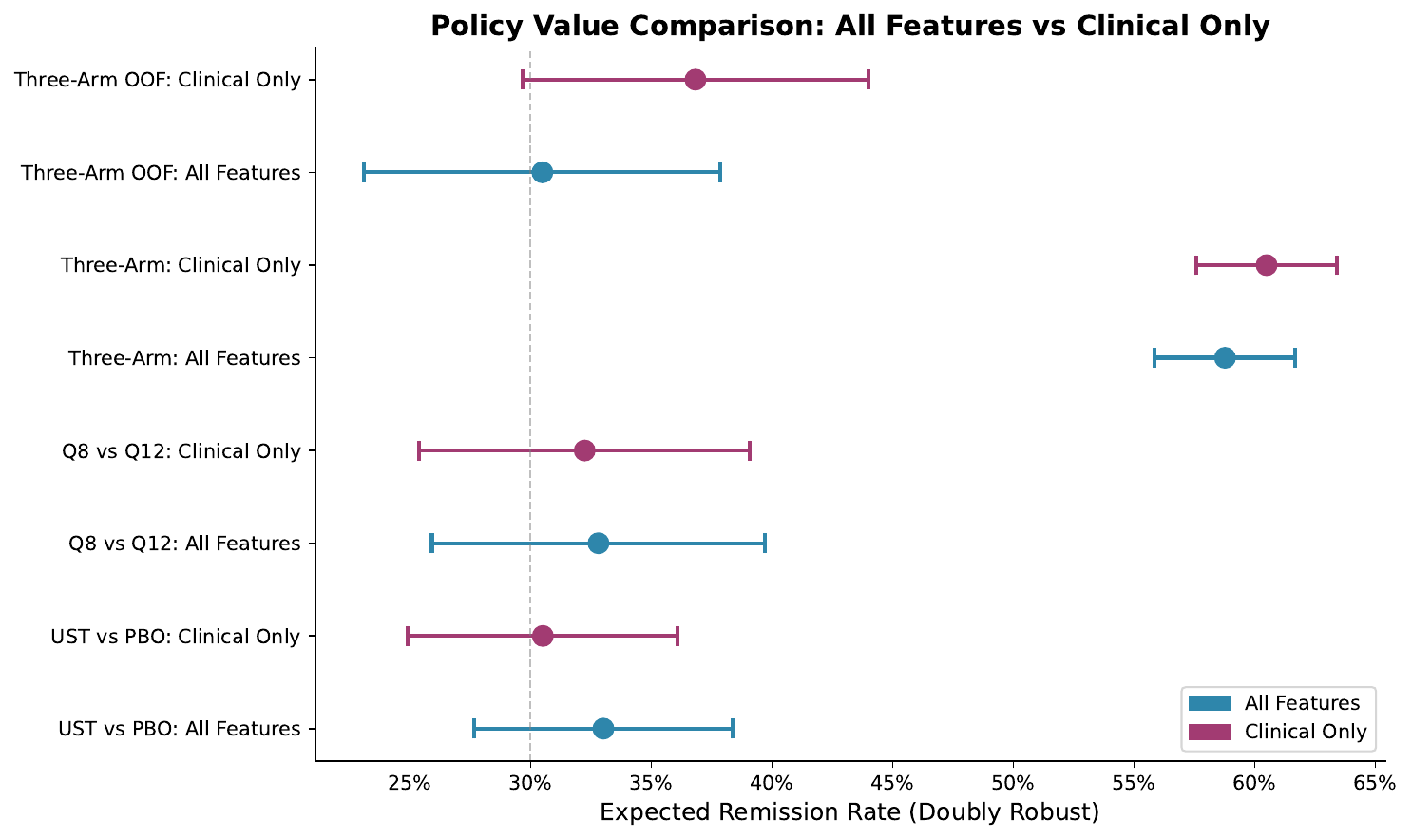}
\caption{Forest plot comparing doubly robust policy values across treatment comparisons. Each point represents the expected remission rate under the learned policy, with 95\% confidence intervals. Blue points indicate policies using all features (including endoscopy); purple points indicate clinical-only policies. The multi-arm out-of-fold comparison shows notably lower performance for all-features policies, suggesting overfitting when endoscopic features are included.}
\label{fig:policy_forest}
\end{figure}

Despite the BLP significance for ustekinumab versus placebo, doubly robust policy evaluation revealed no practical benefit from incorporating endoscopy into treatment assignment rules (Figure~\ref{fig:policy_forest}).

For ustekinumab versus placebo, policies using all features achieved estimated remission rates of 33.0\% (95\% CI: 27.7\%--38.4\%) versus 30.5\% (95\% CI: 24.9\%--36.1\%) for clinical-only policies -- a difference of 2.5 percentage points (95\% CI: $-1.6$ to +6.6 pp) not distinguishable from zero. For dose intensification, the difference was negligible (0.6 pp, 95\% CI: $-4.6$ to +6.0 pp).

In the multi-arm setting, endoscopic features reduced treatment-assignment quality. Out-of-fold evaluation showed lower remission rates for policies using endoscopy (30.5\%, 95\% CI: 23.1\%--37.9\%) compared with clinical-only policies (36.8\%, 95\% CI: 29.7\%--44.0\%), a difference of $-6.3$ percentage points (95\% CI: $-13.3$ to +0.4 pp). The discrepancy between in-sample (58.8\% and 60.5\%) and out-of-fold estimates (30.5\% and 36.8\%) suggests that endoscopic features overfit to noise rather than capturing reproducible treatment-response patterns.

Cross-validated prognostic models provided independent confirmation. In the full cohort (all patients across treatment arms), models with all features attained AUROC of 0.663 and Brier score of 0.204, while clinical-only models achieved superior performance (AUROC 0.684, Brier 0.196; $\Delta$AUROC = $-0.021$, 95\% CI: $-0.054$ to +0.011; Table~\ref{tab:prognostic_performance}). Incremental Brier score analysis confirmed that endoscopic features increased prediction error ($\text{Brier}_{\text{incremental}} = -3.4\%$), consistent with overfitting.

Stratified analyses showed heterogeneous patterns: endoscopic features modestly improved performance in the placebo arm (AUROC: 0.626 vs 0.572, $\Delta$ = +0.055) but degraded performance in the Q8 arm (AUROC 0.634 vs 0.658, $\Delta$ = $-0.025$). This inconsistency reinforces that endoscopic features do not robustly improve outcome prediction.

\begin{table}[ht]
\centering
\caption{Prognostic Performance and Marginal Contribution of Endoscopic Features}
\label{tab:prognostic_performance}
\begin{scriptsize}
\begin{tabular}{lcccc}
\hline
Cohort & Model & AUROC & $\Delta$AUROC & Brier$_{\text{incr}}$ \\
\hline
\textbf{Full cohort} & All features & 0.663 & \multirow{2}{*}{$-0.021$} & \multirow{2}{*}{$-3.4\%$} \\
 & Clinical only & 0.684 &  &  \\
\hline
\textbf{Placebo} & All features & 0.626 & \multirow{2}{*}{$+0.055$} & \multirow{2}{*}{$+7.5\%$} \\
 & Clinical only & 0.572 &  &  \\
\hline
\textbf{Q12} & All features & 0.701 & \multirow{2}{*}{$+0.001$} & \multirow{2}{*}{$+2.0\%$} \\
 & Clinical only & 0.700 &  &  \\
\hline
\textbf{Q8} & All features & 0.634 & \multirow{2}{*}{$-0.025$} & \multirow{2}{*}{$-2.9\%$} \\
 & Clinical only & 0.658 &  &  \\
\hline
\multicolumn{5}{p{0.75\textwidth}}{\vspace{1mm}\footnotesize Note: Full cohort means all three arms combined (placebo + Q12 + Q8, n=561). $\Delta$AUROC = AUROC(all) $-$ AUROC(clinical); positive values indicate endoscopy improves discrimination. Brier$_{\text{incr}}$ = IPA(all) $-$ IPA(clinical); positive values indicate endoscopy improves calibration.} \\
\end{tabular}
\end{scriptsize}
\end{table}

Subgroup ATE analysis validated the CATE importance findings (Table~\ref{tab:subgroup_ate}). The overall ATE for ustekinumab versus placebo was 10.9\% (95\% CI: 2.4\% to 19.0\%). Subgroups defined by top-ranked features showed the expected variation: patients with fecal calprotectin $\geq$150 $\mu$g/g had an ATE of 14.6\% (95\% CI: 5.8\% to 23.0\%) versus 5.4\% for those below threshold; biologic-naive patients showed an ATE of 13.9\% (95\% CI: 0.9\% to 27.3\%) versus 8.0\% for biologic-experienced. The range of subgroup ATEs was 10.9 percentage points (from 3.7\% to 14.6\%), confirming that heterogeneity, while statistically detectable, was modest in absolute terms.

\begin{table}[ht]
\centering
\caption{Subgroup Average Treatment Effects: UST vs Placebo (PBO)}
\label{tab:subgroup_ate}
\begin{scriptsize}
\begin{tabular}{lccccc}
\hline
Subgroup & N & N$_{\text{UST}}$ & N$_{\text{PBO}}$ & ATE & 95\% CI \\
\hline
Overall & 561 & 417 & 144 & 10.9\% & (2.4\%, 19.0\%) \\
\hline
Fecal Calprotectin $<$150 $\mu$g/g & 137 & 98 & 39 & 5.4\% & (-13.1\%, 23.8\%) \\
Fecal Calprotectin $\geq$150 $\mu$g/g & 396 & 298 & 98 & 14.6\% & (5.8\%, 23.0\%) \\
\hline
Age $<$ median & 269 & 204 & 65 & 12.1\% & (-0.6\%, 24.5\%) \\
Age $\geq$ median & 285 & 208 & 77 & 10.1\% & (-1.2\%, 21.0\%) \\
\hline
CRP $<$5 mg/L & 404 & 295 & 109 & 14.0\% & (3.7\%, 23.7\%) \\
CRP $\geq$5 mg/L & 130 & 102 & 28 & 3.7\% & (-13.5\%, 19.3\%) \\
\hline
Biologic-naive & 266 & 199 & 67 & 13.9\% & (0.9\%, 27.3\%) \\
Biologic-experienced & 295 & 218 & 77 & 8.0\% & (-2.6\%, 18.1\%) \\
\hline
pMayo wk8 $<$3 & 349 & 256 & 93 & 14.4\% & (3.2\%, 25.4\%) \\
pMayo wk8 $\geq$3 & 185 & 141 & 44 & 6.4\% & (-5.1\%, 16.9\%) \\
\hline
\multicolumn{6}{p{0.85\textwidth}}{\vspace{1mm}\footnotesize Note: ATEs computed as difference in remission rates (UST $-$ Placebo) within each subgroup. 95\% CIs from bootstrap resampling (5000 iterations). Subgroups defined by clinical thresholds: fecal calprotectin $\geq$150 $\mu$g/g (STRIDE-II \cite{turner2021}), CRP $\geq$5 mg/L (standard inflammation threshold \cite{vermeire2006}), pMayo $\geq$3 (clinical activity \cite{lewis2008}). Age uses median split (no established clinical cutoff).} \\
\end{tabular}
\end{scriptsize}
\end{table}

\section{Discussion}

This case study illustrates a critical distinction in causal machine learning for treatment personalization: the questions which features predict heterogeneity, is the heterogeneity statistically significant, and does it improve treatment decisions can yield contradictory answers. A modular framework that evaluates each question separately -- through permutation importance, BLP testing, and doubly robust policy evaluation -- is necessary to avoid overinterpreting statistical associations as decision-relevant signals.
In the UNIFI maintenance trial, BLP testing identified significant associations between endoscopic features and estimated treatment effects (z = 8.28, p \textless{} 0.001 for ustekinumab versus placebo), yet doubly robust policy evaluation showed no corresponding improvement in expected remission, and out-of-fold multi-arm evaluation showed worse performance when endoscopic features were included (30.5\% vs.\ 36.8\% remission) \cite{steyerberg2019}. This dissociation between heterogeneity detection and decision performance is likely not unique to this trial -- it may arise whenever candidate features carry prognostic information that is mistaken for effect modification.
The diagnostic strategy of comparing a feature set's contribution to outcome prediction (incremental Brier score) against its contribution to policy value across treatment arms proved informative for distinguishing these roles. Endoscopic scores improved outcome prediction in placebo patients (incremental Brier: +7.5\%), consistent with prognostic value, but degraded both prognostic performance in treated cohorts ($-$2.9\% in Q8, $-$3.4\% pooled) and out-of-fold policy value when included for treatment selection. This pattern can be a signature of features that capture disease severity rather than differential treatment response.
Clinical variables such as age, prior treatment exposure, and week-8 inflammatory biomarkers (C-reactive protein, fecal calprotectin) represent stable characteristics with lower measurement error than endoscopic scoring. In this dataset, these routinely available measures were sufficient for the maintenance decisions considered. Subgroup ATE analysis provided model-free validation: features identified as top predictors of heterogeneity corresponded to subgroups with meaningfully different observed treatment effects (e.g., fecal calprotectin $\geq$150 $\mu$g/g: ATE 14.6\% vs.\ 5.4\% below threshold), though the overall range (10.9 percentage points) confirmed modest heterogeneity.

The consistency across analyses -- minimal policy value gains and reduced out-of-fold performance when endoscopic features were included across binary and multi-arm comparisons -- suggests that the limited incremental value of endoscopy is not an artifact of a single modeling choice. Nested cross-validation for the multi-arm policy provided an explicit check on generalization \cite{robins1994,bang2005,kennedy2023}.
More broadly, these results suggest that causal machine learning applications to clinical trials should routinely include policy-level evaluation alongside heterogeneity testing. BLP tests and feature importance analyses are valuable for characterizing the structure of treatment effect heterogeneity, but they do not answer the question that matters most for clinical practice: whether acting on the estimated heterogeneity improves patient outcomes. Out-of-fold doubly robust policy evaluation provides this answer, and the gap between the two may be informative in its own right.
While endoscopic assessment remains central to confirming mucosal healing and informing long-term management \cite{travis2013,shah2016}, our results indicate that week-8 video-derived endoscopic scoring does not improve prediction of week-44 remission beyond routine clinical markers for the decisions studied here. These conclusions do not imply that endoscopic assessment lacks clinical value, but rather that using week-8 Mayo-based summaries for individualized maintenance therapy selection may be challenging given realistic sample sizes and measurement variability. Future work may benefit from reducing measurement error (e.g., standardized reading or automated analysis), incorporating complementary modalities, and testing whether endoscopic information contributes more reliably to other endpoints or decision points.

\section{Conclusion}

We presented a causal machine learning framework for treatment personalization in randomized controlled trials, designed to answer three distinct questions: which features predict treatment effect heterogeneity, whether the heterogeneity is statistically significant, and whether it improves treatment decisions. The framework integrates X-learner estimation, permutation importance, BLP testing, and doubly robust policy evaluation into a modular pipeline applicable across therapeutic areas.
Applied to the UNIFI maintenance trial, the framework revealed a clear dissociation between heterogeneity detection and decision performance. Endoscopic features showed strong statistical associations with treatment effects but degraded out-of-fold policy value when included for treatment assignment. Diagnostic comparison of prognostic contribution against policy value identified endoscopic scores as disease severity markers rather than stable effect modifiers, while routinely available clinical variables captured the decision-relevant variation.
These findings carry a general implication: causal machine learning applications to clinical trials should evaluate candidate features not only for statistical associations with treatment effects, but for their impact on out-of-sample treatment decisions. 
Several limitations warrant consideration. Endoscopic inputs were restricted to Mayo subscoring distributions and the Cumulative Disease Score; other indices or automated features could behave differently. Histology was not available, and the primary endpoint was remission at week 44 rather than time-to-event outcomes. Finally, these results reflect ustekinumab maintenance in UC and may not generalize to other biologic agents or to Crohn's disease.

\backmatter

\section*{List of abbreviations}

ATE - Average Treatment Effect


\noindent AUROC - Area Under the Receiver Operating characteristic Curve

\noindent BLP - Best Linear Predictor

\noindent CATE - Conditional Average Treatment Effect

\noindent CDS - Cumulative Disease Score

\noindent IPA - Index of Prediction Accuracy

\noindent ITT - Intention-to-treat

\noindent MES - Mayo Endoscopic Score

\noindent RCT - Randomized Controlled Trial

\noindent UC - Ulcerative Colitis

\noindent XGBoost - eXtreme Gradient Boosting

\section*{Declarations}

\subsection*{Funding}
Research reported in this publication was supported by the Center for Data-Driven Drug Development and Treatment Assessment (DATA), an industry-university cooperative research center partially supported by the U.S. National Science Foundation under the award number 2209546, and by DATA industry partners.

\subsection*{Conflict of interest/Competing interests}
RS has served as a consultant or on advisory boards for AbbVie, Bristol Myers Squibb, CorEvitas, Eli Lilly, Exact Sciences, Gilead, Janssen, Merck, Pfizer, and Takeda. RS and KN hold intellectual property and equity on medical imaging and endoscopic analysis technologies licensed by the University of Michigan to PreNovo, LLC, AMI, LLC, and PathwaysGI, Inc. The remaining authors disclose no conflicts.

\subsection*{Ethics approval and consent to participate}
The UNIFI study (NCT02407236) was conducted in accordance with the principles of the Declaration of Helsinki and Good Clinical Practice guidelines. The original protocol and all amendments were approved by the institutional review board or independent ethics committee at each participating center, and all participants provided written informed consent prior to enrollment. The present analysis was performed using de-identified data. As this secondary analysis used only anonymized data, additional institutional review board approval and informed consent were not required.


\subsection*{Availability of data and materials}
This analysis uses individual patient data from the UNIFI phase 3 maintenance trial of ustekinumab in ulcerative colitis (NCT02407236), obtained under a data‑sharing agreement with the trial sponsor. The dataset is not publicly available, but qualified researchers can request access from the sponsor according to their data‑sharing policy.

\subsection*{Authors' contributions}
Data acquisition - RS; funding acquisition - CM; project ideation - CM, RS, KN; data preprocessing: RS, CM, ST; software: CM; formal analysis: CM; first draft: CM; final revision: CM, RS, KN.










\end{document}